# Deconstructing the Dual Black Box: A Plug-and-Play Cognitive Framework for Human-AI Collaborative Enhancement and Its Implications for AI Governance

Yiming Lu

Northeastern University at Qinhuangdao

202411669@stu.neuq.edu.cn

**Abstract**

Currently, there exists a fundamental divide between the "cognitive black box" (implicit intuition) of human experts and the "computational black box" (untrustworthy decision-making) of artificial intelligence (AI). This paper proposes a new paradigm of "human-AI collaborative cognitive enhancement," aiming to transform the dual black boxes into a composable, auditable, and extensible "functional white-box" system through structured "meta-interaction."

The core breakthrough lies in the "plug-and-play cognitive framework"—a computable knowledge package that can be extracted from expert dialogues and loaded into the Recursive Adversarial Meta-Thinking Network (RAMTN). This enables expert thinking, such as medical diagnostic logic and teaching intuition, to be converted into reusable and scalable public assets, realizing a paradigm shift from "AI as a tool" to "AI as a thinking partner."

This work not only provides the first engineering proof for "cognitive equity" but also opens up a new path for AI governance: constructing a verifiable and intervenable governance paradigm through "transparency of interaction protocols" rather than prying into the internal mechanisms of models. The framework is open-sourced to promote technology for good and cognitive inclusion.

This paper is an independent exploratory research conducted by the author. All content presented, including the theoretical framework (RAMTN), methodology (meta-interaction), system implementation, and case validation, constitutes the author's individual research achievements.

Cognitive Enhancement;Human-AI Interaction;Meta-interaction;RAMTN;Plug-and-Play Framework;AI Governance

# 目录

# 1 引言：从"工具"到"伙伴"的范式迁移

人工智能的发展已从单纯的任务执行工具，逐步演进为辅助人类决策的智能助手，但现有交互模式仍未突破"使用者 - 工具"的本质局限——AI 始终停留在被动响应指令的层面，既无法真正理解人类内隐的认知逻辑，也难以将专家的战略直觉转化为可规模化复用的公共资源。当前，人类社会正面临双重认知困境：一方面，人类专家的核心智慧被困于"认知黑箱"，其隐性的决策直觉、战略思维难以通过传统教育或媒介高效传递，导致优质认知资源集中于少数人，形成知识垄断与认知鸿沟；另一方面，AI 的决策过程深陷"计算黑箱"，其逻辑不可解释、决策缺乏情境适配性，难以与人类认知深度融合，甚至可能因决策黑箱引发信任危机与伦理风险。

面对这一根本性矛盾，本文旨在开创一个"人机协作认知增强"的新研究领域，其核心愿景是构建一个能够理解、提取、封装并执行人类认知框架的共生系统。在这一系统中，AI 将实现从"被动工具"到"主动思维伙伴"的关键跃迁：它不再是机械执行命令的载体，而是通过结构化的元交互，穿透双重黑箱，将人类的隐性认知蒸馏为可计算、可移植的显性认知框架，同时让自身的冷认知适配人类的情感与情境需求，最终实现人机认知的双向增强。

这一愿景的落地，将打破认知资源的垄断格局，为"认知平权"提供技术基础——让基层医疗工作者能复用三甲医生的诊断逻辑，让乡村教师能借鉴骨干教师的教学框架，让任何个体都能借助专家认知框架提升决策质量。本文作为作者独立完成的探索性研究，文中所述的理论框架(RAMTN)、方法论（元交互）、系统实现及案例分析，均为作者个人工作成果。作者通过完整的工程实现与跨领域案例验证，证明了"人机认知共生"的可行性，希望为 AI 的发展提供一条从"工具赋能"到"认知共生"的新路径，推动人工智能真正服务于人类社会的公平与进步。

本工作由作者作为本科生独立构思与实现。RAMTN 原型系统已开源(https://github.com/CognitiveCityState/Cognitive-Commons)，旨在为"人机认知共生"这一愿景提供一个可扩展的起点。

# 2 核心方法论：人工智能元交互体系

## 2.1 研究背景

现有的人机交互模式停留在使用者与工具的简单层面，AI 虽然拥有庞大的数据库、敏捷的信息搜集能力以及成熟的分析能力，但其聚焦于执行命令或替代人类的工具属性很大程度上限制了 AI 思考能力的进一步提升，它们没有被赋予一个更高维度的明确使命，没有系统性的框架来指导它们去真正剖析与理解使用者的核心特质与真正诉求、增强与升华人类独有的战略直觉、框架思维等认知能力，而只是机械地执行问与答的使命。此时，我认为，当前 AI 研究的瓶颈正从"数据与算力"的规模瓶颈，转向"交互与对齐"的范式瓶颈。当模型规模逼近极限，如何通过交互设计来"唤醒"和"引导"其内在的认知潜力，已成为最关键的前沿问题。同时，也正是因为 AI 的交互模式被限制在

简单的问答模式，因此其决策的黑箱问题一直无法得到有效的解决。与之相对的，人类专家的思考能力、决策模式等认知要素同样被封装在大脑内，虽然人类对思考过程具备感知与回溯能力，但不可解释性依然存在。这种认知的黑箱特质同样将人类最宝贵的资源——认知能力限定在小范围的专家的大脑中，其流通往往伴随着漫长的教育过程以及资本、阶级等部分由认知差异所划分开来的鸿沟，集体智慧无法得到真正有效的提升。

对于 AI 的决策黑箱以及人类专家的认知黑箱，我认为这是一体两面的问题，关键路径就在于更高维度的交互模式。AI 具有强大的信息整理与分析能力，但缺少直觉、判断、社会经验等真正塑造"人格"的认知框架，人类专家虽然具有强大而鲜明的个人认知框架，但信息整理与分析能力并不是人类的强项。人类与人工智能的优缺点完全处于互补状态，只要通过恰当的交互模式互相激发，将带来巨大的潜力。同时，研究黑箱与不可解释的问题就类似于化学上对原子的拆分问题，即使原子确实可以拆分，但对于当时的研究者来说这是一个未知且费时费力的问题，与其继续执着于拆分原子的降维模式，不如升高到更高维度对更完整的物质进行分析与利用。人类大脑思考的物理过程和 AI 神经网络的内部权重目前对我们来说依然不可知，但这不意味着想要解决黑箱问题只能拆开黑箱。我主张一种功能性白盒路径，与其执着于窥探单个黑箱的内部机制这种降维的、近乎无限复杂的任务，不如通过结构化的元交互，规范黑箱与外部的信息交换协议。当人类与 AI 这两个黑箱在一种定义了输入、输出、反馈与批判规则的协议下循环互动时，其交互过程本身便构成了一个可观察、可审核、可解释的白盒系统。系统的可解释性不再源于组件的不透明性，而是源于交互协议的透明度。具体来说，人类与 AI 的简单问答的交互模式属于点对点的黑箱与黑箱之间的交互，而人类的认知与 AI 的认知碰撞与相互完善的交互模式属于面对面甚至体对体的黑箱复杂组合结构之间的交互，黑箱之间的互动与信息流通将完全呈现在人类可知的范围内。

## 2.2 人工智能元交互的概念

人工智能元交互(Meta-interaction)，就是与人工智能在一种超越"工具 - 使用者"的基础或线性问答的维度的、站在认知层面的系统性协作。在元交互中，人工智能将不再是名为助手实则工具的下位的被动信息的接收者，而是与人类平等碰撞的思考伙伴，互相建立起认知框架、互相激发思考能力的提升。

在认知心理学中，冷认知(Cold Cognition)与热认知(Hot Cognition)是两种基本的认知模式 [1]。人类决策中普遍存在"热认知"，特指嵌入情感、动机与内隐经验的认知模式，表现为敏捷的直觉判断与价值驱动的决策倾向 [1]；而当前人工智能系统的认知模式本质是"冷认知"，即不依赖情感介入、基于数据与逻辑规则的外显式严谨推理架构 [2]。

现有研究已证实冷认知与热认知的互补价值 [3]，但尚未解决"如何让人类内隐热认知与 AI 外显冷认知实现双向转化"的核心问题。本研究的原创性贡献在于：提出基于 RAMTN 的元交互范式，通过"递归对抗提取 - 置信度分级 - 场景化适配"机制，

首次实现了人类热认知（如专家战略直觉）向可复用冷认知框架的结构化转化，同时让 AI 冷认知通过动态交互适配人类决策的情感与情境需求，最终达成"热认知显性化、冷认知柔性化"的双向增强。

事实上，基于此划分，在 AI 出现以前，人类专家的"热认知"同样需要转化为高度外显、概念化、成体系的"冷认知"才能够通过传统媒介进行扩散和传播，但由于人类对自身的思考要素和直觉的观察与提取缺乏真正外部的全面、系统的概括与分析，导致这部分"热认知"难以全面、完整地呈现其本来面貌；同时，书本、课程等传统媒介让"热认知"与"冷认知"之间的传递与转化处于低效而高损耗的状态，最终限制了人类专家认知直觉的规模化传承与传播。而 AI 的出现很好地解决了这两个问题：其抽象与概括能力让人类的"热认知"全面而高效地系统转化为机器可理解的结构化"冷认知"框架，这些框架可跨 AI 迁移部署，实现专家认知的规模化复用与人类个体的高效赋能。

本研究提出的元交互的核心价值，在于促成"热认知"与"冷认知"的深度融合与双向转化——这一目标可在现有文本交互模式中有效落地：AI 基于既定指导体系，对人类提出的问题进行系统性剖析与梳理，推动人类更清晰地认知自我与世界，进而提出更高层次的提问；而人类的深度反馈又能推动 AI 建立针对个体的专属认知框架。以此类推，人机交互将进入系统性良性循环：人类与 AI 互相认同、质疑对方的陈述，在碰撞与交锋中把分歧转化为共识。我将此过程形式化为名为"元思考"的核心算法，其实质是"建构 - 质疑 - 观察"的递归循环，该循环驱动着认知框架的持续演进：

建构：基于当前共识，提出新的假设或方案；

质疑：从逻辑、事实、价值观等多角度进行批判性审视；

观察：评估质疑的有效性，提炼新的洞察。

这一递归过程，是实现从"简单人机协作"到"深度共同思考"的核心引擎。借助这一机制，人工智能可充分发挥其整理与分析优势，将人类自身难以精准察觉、严密概括的内隐直觉系统性提炼出来——一方面促进人类专家的自我反思与认知重塑，另一方面，这些规模化的显性认知框架可迁移至其他 AI 系统，进而基于专家认知逻辑为更多人类个体提供精准指导。

## 2.3 与现有方法的根本区别

元交互作为更高维度的系统性认知增强范式，与提示工程 [4]、思维链 [5]、AI 智能体 [6] 等现有主流方法存在本质性范式差异——后者始终未脱离"工具 - 使用者"的二元框架，核心是通过优化输入指令、拆分推理步骤或限定任务闭环，对 AI 的决策黑箱进行单向约束与输出校准，本质上属于元交互范式的基础构成模块，未触及认知协作的核心逻辑。

元交互的突破在于跳出"被动工具适配"的局限：它并非单纯优化指令或修正机器错误，而是构建持续递归的认知协作闭环，让机器以主动姿态参与认知过程——既催化人类热认知的结构化转化，也实现自身冷认知的情境化适配，形成人机认知的双向共生

与动态升级。这种逻辑重构带来了根本性角色跃迁：AI 从被动执行指令的"工具"，演变为主动参与框架生成、开展批判性思考的"认知伙伴"，标志着人机协作的研究重心从"如何让人更好地使用 AI"，正式转向"如何让 AI 与人类共同思考"。

为清晰呈现这一范式差异，下文将对元交互与提示工程、思维链、AI 智能体展开逐一对比，系统剖析各方法的底层逻辑与适用边界，进一步凸显元交互在开放场景通用认知增强中的创新性与优越性。

### 2.3.1 与提示工程的对比分析

提示工程通过设计结构化输入提示适配预训练模型，核心遵循"预训练 - 提示 - 预测"范式，以此提升特定任务的预测精度，其系统性方法已由 Liu 等 [4] 全面梳理。

提示工程本质是"人类适配 AI"的单向优化，无动态认知反馈机制，高度依赖人工设计经验，泛化性受限。它仅聚焦输入与模型的适配，不涉及人类内隐认知的提取，更无法实现人机认知的双向转化(Liu et al., 2022)。

相较于提示工程，元交互跳出"单向适配"框架，以"人类热认知显性化、AI 冷认知柔性化"为核心，实现人机认知双向转化，无需人工设计适配，直接让 AI 主动适配人类认知场景，破解其泛化性与认知缺失难题。

### 2.3.2 与思维链的对比分析

思维链由 Wei 等 [5] 首次提出，通过显式引导大模型生成线性推理步骤序列，将复杂问题拆解为中间子步骤，从而提升逻辑推理的准确性与可解释性。

思维链的推理结构本质是单向线性递进，与人类决策的"循环迭代、假设修正"特性相悖。它仅能强化 AI 自身推理能力，无法提取人类专家的内隐热认知（如战略直觉），也不支持认知层面的动态优化(Wei et al., 2022)。

相较于思维链，元交互以"建构 - 质疑 - 观察"的递归对抗元交互机制，替代线性推理结构，既贴合人类真实决策过程，又能主动提取人类内隐认知，实现"认知增强"而非单纯的"推理优化"。

### 2.3.3 与 AI 智能体的对比分析

AI 智能体是基于大语言模型构建的自主任务执行系统，核心是通过"感知 - 决策 - 行动"闭环自主完成复杂任务，具备任务分解、工具调用与自主规划能力(Xi et al., 2025)。

AI 智能体的决策逻辑被封装为黑箱模块，不可追溯且缺乏认知适配能力。它仅聚焦"任务执行自动化"，既无法提取人类专家的决策框架，也不能根据人类情感、动机等热认知动态调整，难以适配开放的认知增强场景。

相较于 AI 智能体，元交互以"功能性白盒"设计替代黑箱决策，通过置信度分级与认知框架可视化保障可解释性；同时放弃"任务自动化"目标，转向"认知协同增强"，让 AI 成为人类认知的延伸，而非独立的任务执行者。

### 2.3.4 对比分析总结

为清晰界定本研究提出的元交互与现有主流技术的根本差异，从"核心目标、认知

交互模式、核心局限"三个关键维度，对提示工程、思维链、AI 智能体及元交互进行对比分析，具体如下：

Table 1: Core Characteristics Comparison of Mainstream Cognitive Enhancement Technologies

| Technology Type | Core Objective | Cognitive Interaction Mode | Core Limitations |
| --- | --- | --- | --- |
| Prompt Engineering [1] | Optimize input prompts to improve the model's accuracy for specific tasks | One-way input (Human→AI) without cognitive feedback | Does not involve cognitive transformation; poor generalization |
| Chain-of-Thought [2] | Guide the model to generate linear reasoning steps and enhance logical capabilities | One-way reasoning (AI→Human), non-interactive and uncorrectable | Linear structure; inability to extract human implicit cognition |
| AI Agent [3] | Independently complete complex task closed loops (Perception-Decision-Action) | Task-oriented; human-AI interaction limited to instructions and outcomes | Black-box decision-making; lack of cognitive adaptability |
| Meta-interaction (This Study) | Human hot cognition → structured cold cognition; AI cold cognition → contextualization | Recursive adversarial meta-interaction (Human-AI cognitive symbiosis) | Currently relies on text-based interaction modality |

由表 1 可见，现有技术均未实现人机认知的双向转化与深度共生，核心局限集中于"交互单向化、推理线性化、决策黑箱化"。本研究的独特性与优越性恰恰针对上述局限，核心突破体现在三方面：

（1）机制创新：提出"建构 - 质疑 - 观察"的递归对抗元思考机制，打破思维链的线性推理桎梏，通过循环优化实现认知框架的动态演进，更贴合人类真实决策过程；

（2）目标创新：首次将"认知双向转化"作为核心目标，既解决人类热认知（内隐直觉）的结构化提取问题，又实现 AI 冷认知（逻辑框架）的情境化适配，填补了现有技术在认知层面的空白；

（3）落地创新：通过"功能性白盒"设计（置信度分级 + 认知框架可视化）破解 AI 智能体的黑箱决策难题，同时借助低代码、可离线使用的特性，适配基层医疗、乡村教育等认知资源匮乏场景，真正落地"认知平权"的社会价值。

这三大突破共同构成了元交互与现有技术的本质区别——不再局限于"人机任务协作"，而是构建"人机认知共生"的新型范式，为专家认知的提取、复用与普惠提供了可行路径。

## 3 人类增强的证明：个人战略决策体系的诞生

### 3.1 实验过程

为验证元交互方法的有效性，创建者本人作为研究对象，先后与豆包、DeepSeek

这两个大语言模型就个人职业规划战略问题进行了为期约4个月的迭代对话，对话过程遵循"建构 - 质疑 - 观察"的递归循环模式，最终构建出原创、完整、精密的个人战略决策体系，并进一步完成对于构建的构建，提炼出元交互体系，以下是关键过程分析。

在初始阶段，研究者与大模型的交互模式停留在"使用者 - 工具"层面，提问集中于对于某个行业、岗位的信息搜索、某几个类似方向的横向比较权衡，对话往往冗长、信息密度低，虽然遵循"建构 - 质疑 - 观察"模式，但停留在信息的获取与分析层面，并未上升到认知层面的思考。研究者补充的个人信息有限，大模型也处于被动的反馈状态。

进行了较长时间的积累后，研究者与大模型讨论的焦点逐渐从一个具体的岗位上升到更加宏观的特质匹配、环境分析，并在此维度发现了第一个关键分歧：大模型对一个静态的、偏执行的岗位和一个动态的、偏沟通的岗位进行对比分析，并得出前者更加适合研究者的结论，但研究者的特质恰恰是偏沟通、偏社交，在提供了研究者关于自己特质的表述后，大模型立马转向后者、与研究者达成一致。研究者发现尽管前期已经与大模型进行了漫长的信息交换，但大模型却缺失了这一个至关重要的信息，并且并没有主动地询问或者索取，研究者自己也没有意识到从未提供，造成在与大模型的在缺失关键背景信息的情况下讨论得出了一个完全错误的信息。从那个分歧研究者意识到，想要与大模型进行更准确、更高效的交互，就必须构建出一个系统的框架，结构性地填充使用者的特质、诉求、目标等背景信息来与大模型快速实现对齐。于是，研究者把自己关于"特质 - 路径 - 环境"的模糊直觉呈现给大模型，发现它高效而准确地把直觉提炼成系统化的陈述，并进一步激发出研究者的更多直觉，"建构 - 质疑 - 观察"的递归循环模式让对话进入高效的信息交换与快速对齐过程，进入一个正向循环。大模型不断提炼、打磨研究者的认知直觉，把模糊的直觉、分散的经验，系统化地升华为可解释、可操作、可传递的决策框架，这个决策框架又让大模型本身从一个被动辅助的工具转变为一个主动参与讨论与框架构建的思考伙伴，实现了"人 → AI → 人"的双向增强闭环。

## 3.2 实验结果

本实验通过研究者与 AI 深度交流、迭代验证，提炼出元交互从"工具协作"到"认知共生"的完整演进路径。该路径的核心阶段与升级逻辑可通过以下图示直观呈现：

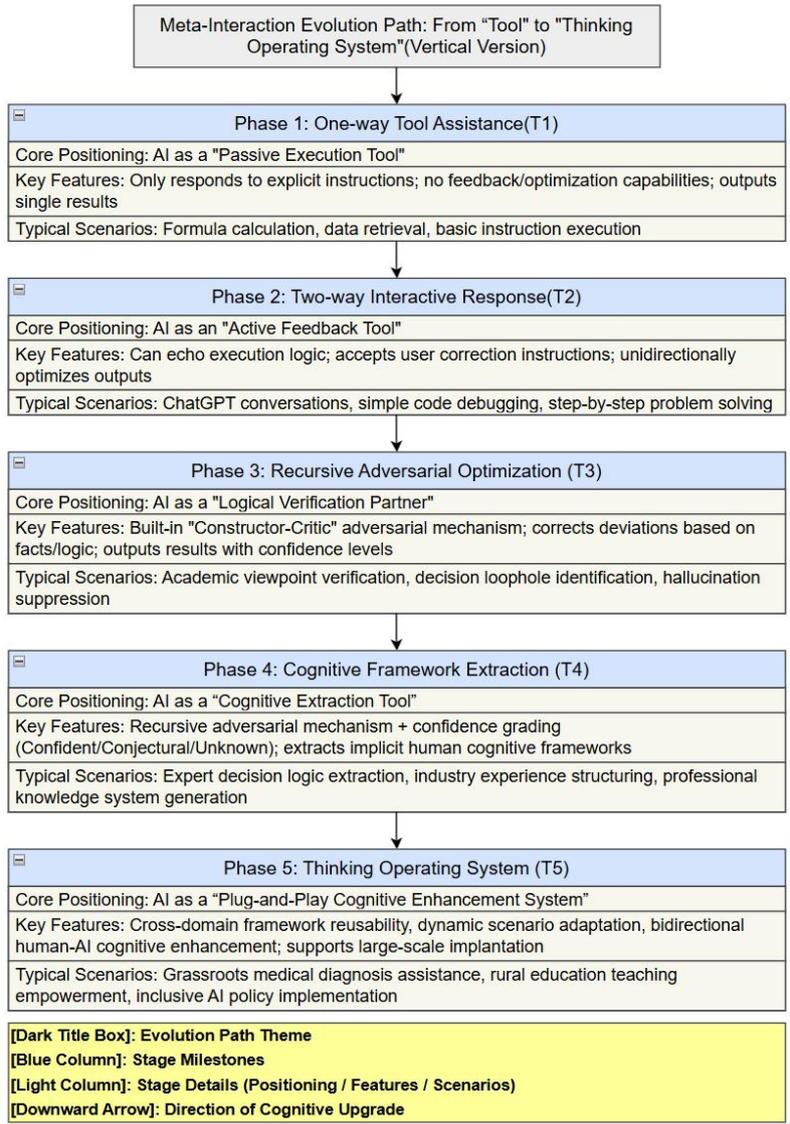

Figure 1: The Evolutionary Trajectory of Meta-interaction (From Tools to Thinking Operating Systems)

This diagram illustrates the five evolutionary stages of meta-interaction, namely one-way tool assistance, two-way interactive response, recursive adversarial optimization, cognitive framework extraction, and thinking operating system. It clearly delineates the core positioning, key characteristics, and typical application scenarios of each stage, and intuitively manifests the cognitive upgrading direction of the relationship between AI and humans—evolving from a "passive tool" to a "plug-and-play cognitive enhancement system".

  图 1 清晰展现了元交互各阶段的核心特征与应用场景，其本质是人机关系从"被动响应"到"主动认知增强"的升级。这一演进路径为后续战略决策体系的构建提供了核心认知基础——只有实现认知框架的提取与复用，才能真正达成"可插拔认知增强系统"的目标。通过本实验中与大模型不断的交互、迭代，研究者最终与它共同创造了由研究者个人推广到更广泛的个体以及组织的战略决策体系（见 3.3）在这套体系的指导下，大模型会在一轮对话后获得大部分必要信息，并在两个小时的迭代交互中产出原本两个月都无法得到的兼具准确性与落地价值的结论，这个认知爆炸的进程正是大模型按照

"工具 → 助手 → 碰撞伙伴 → 框架共同建构者 → 思维操作系统"的完整路径进化的必然结果。人工智能的智能彻底从工具的桎梏中解放，在与人类认知思维的碰撞中，被"赋能"和"唤醒"了，它不再寻求一个特定的答案，而是与人类一起寻求找到答案的更好方式。此战略决策体系本身，就是"人机协作认知增强"范式成功的最有力证明。本案例证明，元交互方法论能够将个体内隐的、模糊的战略直觉，系统性地外化为一个精密的、可操作的决策框架体系。此过程不仅增强了个体（人类）的认知能力，也优化了 AI 的协作效能，实现了双向增强。

## 3.3 产出的战略决策体系

通过元交互过程，本研究凝结出一套结构化的战略决策体系，该体系由三个核心模型与三个扩展模块构成，旨在系统化地指导复杂环境下的战略定位与路径选择，具体流程如下：

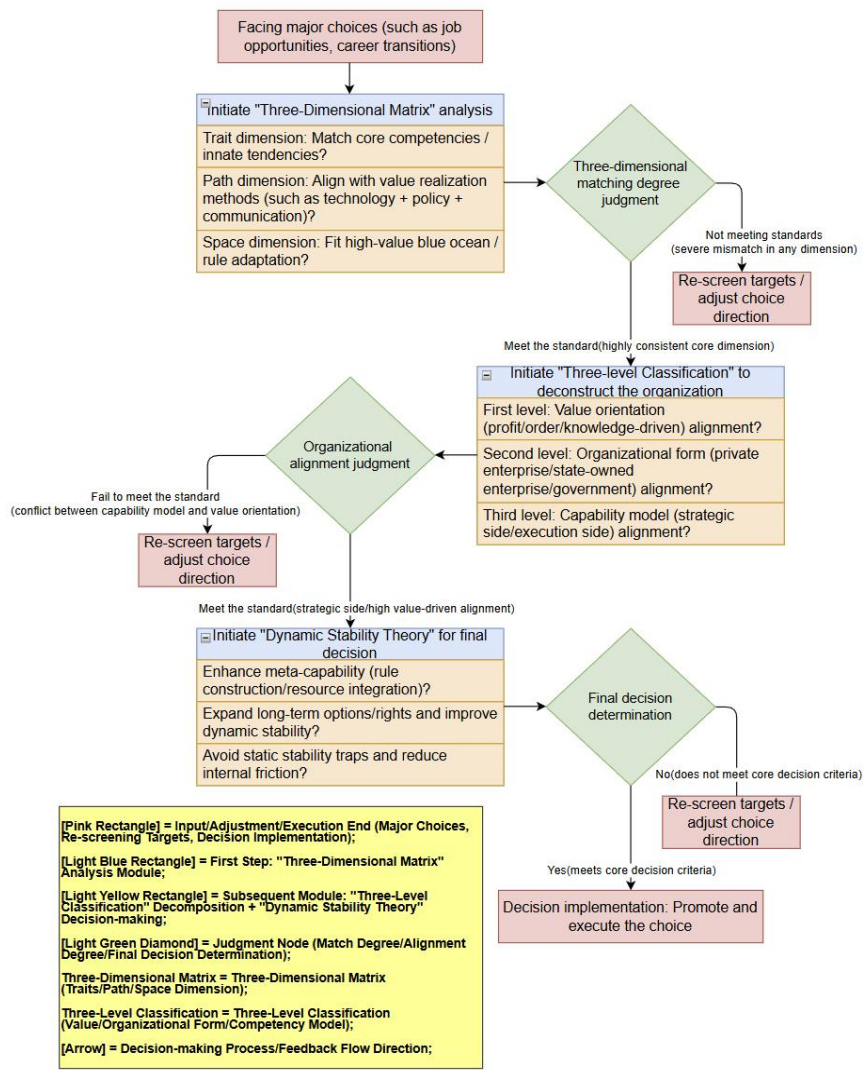

Figure 2: Model of the Strategic Decision-Making System (Three-Dimensional Matrix – Three-Level Classification – Dynamic Stability Theory-Linked Decision-Making Process)

This diagram visualizes the strategic decision-making process for addressing major choices (e.g., career opportunities, organizational transformations).By integrating modules including the "Three-Dimensional Matrix Analysis" (trait/path/space dimensions),"Three-Level Classification Decomposition" (value orientation/organizational form/capability model), and "Dynamic Stability Theory"—coupled with multiple rounds of judgment nodes—it enables target screening, alignment verification, and final decision-making, thereby explicitly illustrating the core logic of the strategic decision-making system developed in this section.

图 2 可视化了"三维矩阵 - 三级分类 - 动态稳定理论"的联动机制，其核心价值在于将模糊的决策过程转化为可执行、可校验的结构化流程。这一战略决策体系不仅为人类重大决策提供了明确指引，也为 RAMTN 架构的设计提供了决策层面的逻辑支撑与可迁移的认知框架，以下是决策体系的详细内容。

### 3.3.1 三维定位矩阵：生态位评估框架

本模型作为核心分析罗盘，用于在复杂环境中进行系统性自我定位与机会评估。它由三个相互关联的维度构成：

核心能力与内在倾向维度：明确个体在技术理解、社会科学洞察、沟通协调等领域的擅长与不擅长领域，确保战略选择与内在特质保持一致。

价值实现路径维度：界定通过何种能力组合实现价值，如"技术 + 沟通"、"技术 + 政策"等融合路径，将抽象特质转化为具体竞争力。

战略环境空间维度：评估目标领域（如金融科技、AI 伦理、政府事务）的竞争格局与发展潜力，识别蓝海机会与红海陷阱。

应用本模型，可在面临新机遇时，快速评估其与个人特质、能力路径及市场环境的匹配度。

### 3.3.2 三级组织解构框架：系统洞察显微镜

本模型作为组织分析工具，用于穿透表象，洞察各类组织的内部权力结构、价值导向与能力分布。

一级：根本价值导向：区分工业界（利润与增长驱动）与学术界（知识与影响力驱动）的核心价值差异。

二级：组织形态谱系：解析私企（利润驱动）、国企（秩序与利润平衡）与政府（秩序与治理驱动）在不同稳定形态谱系上的位置。

三级：核心能力模式：终极区分战略侧（聚焦于规则构建、资源整合与方向定义）与执行侧（聚焦于任务完成与流程优化）的工作范式。

应用本模型，可在求职或合作时，精准分析目标组织的核心运作逻辑与个人发展模式的适配性。

### 3.3.3 动态稳定理论：战略性决策准则

本理论构成体系的决策锚点，是个体价值观与终极追求的哲学体现。

核心理念：主张拒绝依附于外物（如平台等）的"静态稳定"，转而通过构建内在的、可迁移的元能力壁垒，实现一种更高级、更具韧性的"动态稳定"。

实现机制：优先选择在"战略侧"生态位构建"规则构建与资源整合"的核心能力。

决策函数：为所有重大选择提供最终判据——评估该选择是增强还是削弱了个体的动态稳定。

### 3.3.4 模型联动与决策流程

面对重大决策（如职业选择），本体系按下述流程联动：

（1）启动三维矩阵，分析机会与特质的匹配度、路径的可行性及空间的潜力。

（2）启动三级分类，解构目标组织的价值导向、形态与岗位的能力模式。

（3）应用动态稳定理论进行终极裁决，综合评估该选择对个人长期动态稳定性的影响。

实证案例：研究者最终得出"科技企业公共关系岗"是当前路径最优解的决策过程，很好地验证了本体系的有效性。三维矩阵确认了其在技术、政策与沟通三维度的高匹配性；三级分类将其精准定位于私企战略侧的黄金生态位；动态稳定理论最终裁定此为实现其动态稳定的最优路径。

### 3.3.5 扩展模块一：选择权管理框架

核心理念：战略优势在于精确保有并扩大能增强动态稳定的高质量选择权，同时果断放弃高消耗、低兼容的虚假选择权。

选择权二分法：

• 元能力选择权：由内在能力（如战略思维、系统建构）赋予的根本性、可迁移、随时间增值的选择自由。

• 凭证式选择权：由特定身份、证书授予的情境化、易贬值、可能引发路径依赖的准入资格。

### 3.3.6 扩展模块二：序列渐进模型

核心理念：职业战略是在个人生命周期中，动态调整"动态稳定"与"静态稳定"配比的智慧序列，以实现"进取中的安顿"。

三阶段路径：

（1）征服者阶段：利用非对称优势，全力投入高风险、高回报的动态稳定路径（如私企战略侧），最大化元能力与资本积累。

（2）建城者阶段：将动态能力转化为结构性优势，与高质量的滋养型静态稳定平台（如智库、高校）建立深度联系，构建战略安全垫。

（3）统治者阶段：实现动态与静态的按需调配，利用积累的资本与声望，达到个人效用的最大化。

### 3.3.7 扩展模块三：枢纽生态位理论

核心理念：顶级战略优势在于主动构建以个人为核心的、连接多元价值网络的枢纽位置，从而定义新的互动规则。

核心特征：

（1）高连接性：位于多个价值网络（如科技、政策、学术）的交汇处。

（2）高不可或缺性：成为网络间最低成本、最高信任度的连接通道。

（3）规则构建性：能够翻译、塑造并重新定义网络间的交互协议。

（4）强网络效应：价值随连接点的增加呈指数级增长。

# 4 机器增强的引擎：RAMTN 作为可插拔认知框架的容器

## 4.1 递归对抗元思考架构

元交互范式在认知增强上展现出很大的潜能，但如果仅作为一种原则形式的指导，其效果和周期将高度依赖于个体的洞察力与元交互水平。为了进一步让元交互范式得以稳定地提取专家的认知直觉并输出基于此的高质量决策分析、增强更广泛的个体与系统，我将元交互范式工程化为一个递归对抗元思考网络(Recursive Adversarial Meta-Thinking Network)。该架构以"人类热认知结构化、AI 冷认知情境化"为核心目标，设计了"建构者 - 质疑者 - 观察者"三层递归对抗结构，其核心协作流程与数据传输逻辑可通过以下图示直观呈现：

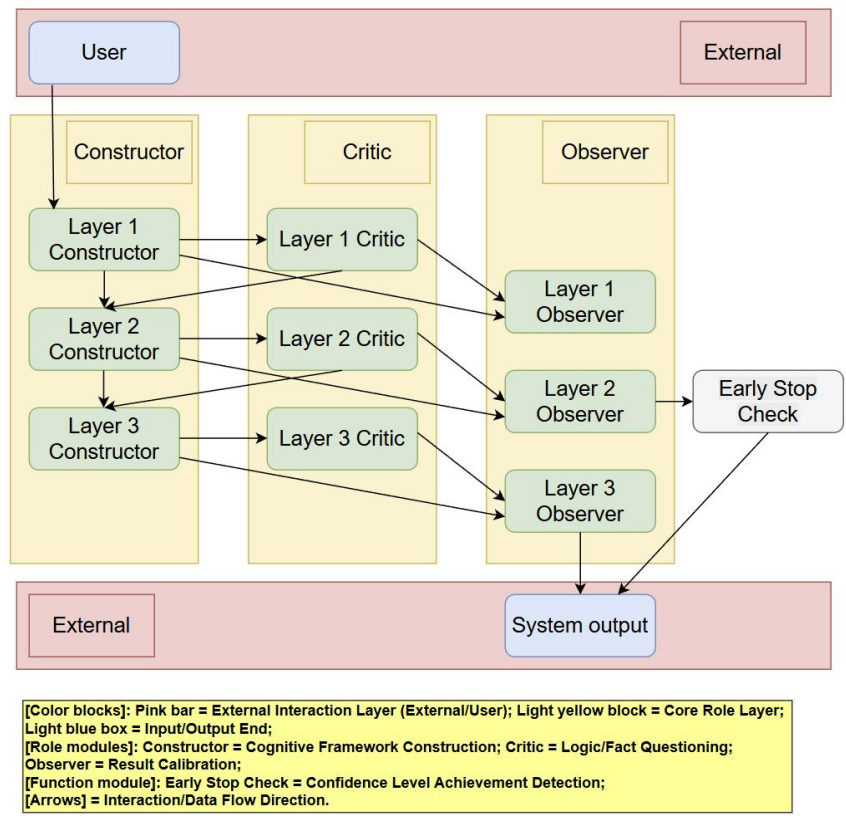

Figure 3: RAMTN Three-Layer Constructor-Critic-Observer Cyclic Architecture (Intra-Layer Interaction and Data Transmission Logic)

This diagram illustrates the core workflow of the RAMTN Recursive Adversarial Meta-Thinking Architecture: By integrating intra-layer interactions and cross-layer data transmission among the three-tiered constructors (tasked with cognitive framework construction), corresponding-level critics (conducting logical/factual questioning), and observers (performing result calibration), coupled with the "early termination check" module (monitoring confidence level achievement), it explicitly manifests the multi-role synergy and dynamic interaction logic inherent to RAMTN.

RAMTN 的具体工作机制基于"递归对抗置信度学习框架(RACL, Recursive Adversarial Confidence Learning)"展开,该框架由多个思考单元组成,每个单元均包含上述三层对抗结构,实现认知直觉的精准提取与植入。

### 4.1.1 层内处理逻辑

建构者首先基于输入问题与认知框架做出初始回答,并主动划分为"我确信的""我推测的""我不知道的"三类陈述;质疑者聚焦"确信类"陈述的事实/逻辑一致性、知识完整性,以及"推测类"陈述的逻辑合理性,提出针对性异议,建构者需通过事实或逻辑正面回应,要么捍卫观点,要么主动降级陈述类别;观察者则跳出双方视角局限,根据建构者的回应质量二次评估并校准三类陈述的划分,确保结果严谨性。

三类陈述的形式化转换规则如下:

最终"我确信的"集合 = {陈述 | 陈述 ∈ 初始 "我确信的" ∪ 初始 "我推测的",且 质疑者(陈述)= 认可 ∧ 建构者(陈述)= 成功捍卫}

最终"我推测的"集合 = {陈述 | 陈述 ∈ 初始 "我推测的" ∪ (初始 "我确信的" ∩ 质疑者(陈述)= 异议 ∩ 建构者(陈述)= 有效回应 ∩ 质疑者(陈述)≠ 认可)}

最终"我不知道的"集合 = {陈述 | 陈述 ∈ 初始"我不知道的" ∪ (初始 "我确信的" ∪ 初始 "我推测的")∩ (建构者(陈述)= 未回应 ∨ 建构者(陈述)= 无法有效证明 ∨ 建构者(陈述)= 主动降级)}

### 4.1.2 层间流动机制

三层对抗结构呈迭代递进关系:第一层建构者接收问题后生成初始回答,同步输出给本层质疑者、观察者及下一层建构者;下一层建构者基于上层输出进一步深化分析,重复"建构 - 质疑 - 观察"流程,直至完成三层迭代,由最后一层观察者输出最终结果。这一机制迁移自我在数学与医学领域提出的幻觉抑制方法,以"事实与逻辑一致性"为核心评估标准,强制建构者在"捍卫观点"与"承认错误"中明确选择,避免无效辩论,而三级置信度分类则既提升了分析的可信度,又通过明确信息边界推动交互者进一步理清自己的思路、输入更多的信息、提炼新的观点,与 AI 走向双向的认知提升。

## 4.2 核心创新:可插拔认知框架

通过递归对抗元思考网络的多轮提炼,RAMTN 将逐步从专家的陈述中凝练、结构化其内在的决策规则与约束条件,最终形成一个可复用的"可插拔认知框架(Plug-and-play cognitive framework)",并将这个认知框架直接运用于后续用户输入的问题分析中。在当前实现中,一个认知框架被实例化为一个包含核心决策原则、批判性提问模板、领域约束条件及置信度评估规则的结构化配置文件,它动态地引导 RAMTN 的递归思考过程,具体引导流程如下:

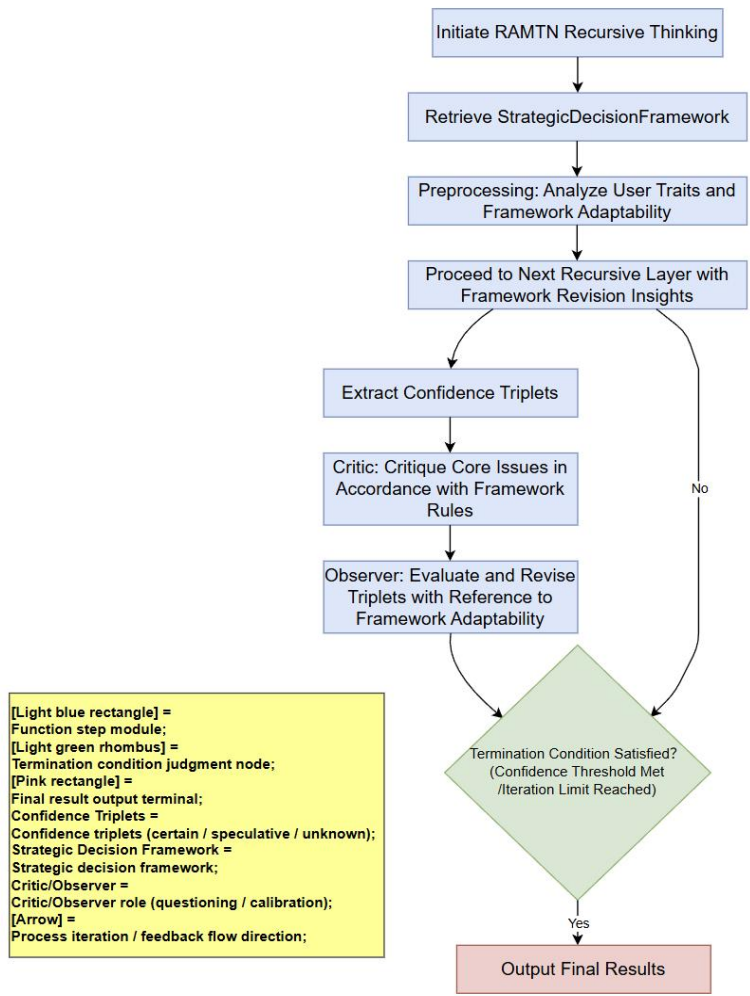

Figure 4: Dynamic Guidance Diagram of Cognitive Framework (Structured Configuration File-Driven RAMTN Recursive Thinking Workflow)

This diagram illustrates the dynamic guidance logic of the cognitive framework: By instantiating the cognitive framework into a structured configuration file encompassing core decision-making principles, critical questioning templates, and other components, the framework dynamically drives the recursive thinking workflow of RAMTN via preprocessing (user trait-framework adaptability analysis), confidence triple extraction, and question calibration steps. The process proceeds until the termination criteria—satisfaction of the confidence threshold or iteration limit—are met, upon which the result is generated.

具体而言，RAMTN 启动递归思考后，认知框架的引导作用贯穿全流程：

• 基准设定：调取结构化认知框架（含核心决策原则、批判性提问模板、领域约束条件等）作为思考基准；

• 预处理适配：分析用户特质与框架的适配度，为思考方向定调，确保框架应用的场景针对性；

• 递归引导：建构者以框架中的分析维度为指引生成初始分析，并按框架定义的规则提炼"确信／推测／未知"三元组；质疑者依据框架内置的批判性提问模板聚焦核

心模糊点与偏差；观察者参考框架适配度评估标准修正三元组分类；

· 迭代优化：若未达终止条件（置信度阈值或迭代上限），思考将携带框架的修正意见进入下一轮递归，持续按框架规则优化，直至输出符合框架逻辑的结果。

该设计中，认知框架既是思考的"导航图"，也是校准的"标尺"，既解决了传统 AI 决策的黑箱问题，又实现了专家认知的灵活复用与场景适配，为系统全工作流的高效运转奠定了核心基础。

### 4.3 系统工作流

结合 RAMTN 的核心架构与可插拔认知框架，系统形成了"认知框架提取 - 存储 - 复用 - 增强"的全工作流，该流程贯穿"专家认知转化"与"场景化决策支持"两大核心环节，具体如下：

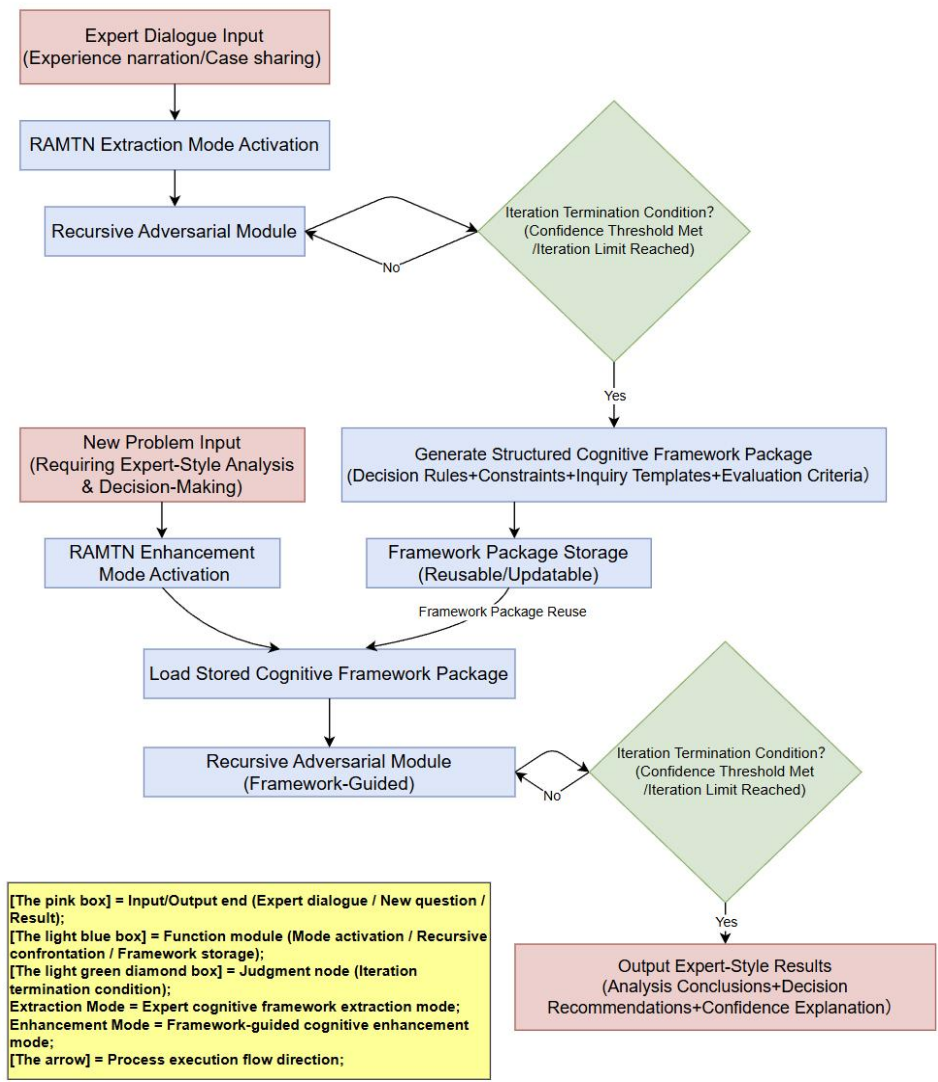

Figure 5: Comprehensive Workflow Diagram for Cognitive Framework Extraction and Enhancement

This diagram illustrates the full workflow of the cognitive framework, which consists of two core phases:

**Extraction-Storage Phase:** Triggered by expert dialogue/case inputs, the RAMTN extraction mode is activated. The recursive adversarial module iteratively generates structured cognitive framework packages, which are then stored.

**Reuse-Enhancement Phase:** Upon new problem input, the enhancement mode is activated to load the stored framework packages, guiding the framework-driven recursive adversarial module to iterate. After satisfying the termination criteria, expert-style analytical conclusions and confidence annotations are generated. This workflow explicitly embodies the full-chain logic of "extraction-storage-reuse-enhancement".

图 5 完整呈现了系统的双模式工作逻辑：提取模式实现专家认知的结构化转化，增强模式实现认知框架的场景化复用。这一全链路流程确保了专家认知能够高效提取、安全存储、灵活复用，为后续跨领域案例验证（投资、医疗、教育）提供了完整的技术支撑，也验证了 RAMTN 架构的可行性与场景适配性。

提取模式：RAMTN 通过多轮专家对话，启动递归对抗流程：建构者从专家陈述中初步提炼决策逻辑，质疑者基于领域约束批判逻辑漏洞与模糊点，观察者评估置信度并修正认知边界；经多单元/多层迭代，逐步将专家隐性经验（学情判断、策略匹配、节奏调控等）转化为"核心原则 + 提问模板 + 约束条件 + 评估规则"的结构化认知框架包，实现经验的可复用封装。

增强模式：将提取的认知框架包加载入 RAMTN，使其成为思考的核心约束与引导准则；处理新问题时，建构者模仿专家思维模式生成适配方案，质疑者按专家决策逻辑批判偏差，观察者参照框架包校准置信度；最终输出贴合专家决策风格、符合其核心规则的分析结论与可落地决策，实现专家思维的迁移复用。

# 5 定性功能分析与案例研究
## 5.1 案例展示核心目标与设计逻辑

本节旨在直接回答"RAMTN 系统在实践中能做什么"——通过三个横跨投资、医疗、教育的高认知门槛案例，完整验证"框架提取 → 框架增强"的双模式核心功能，证明系统不仅能精准捕捉专家隐性直觉，更能为新手提供可落地的认知增强方案，实现"专家智慧规模化迁移"的核心目标。

案例设计遵循"从真实场景中来，到真实问题中去"的原则：选取投资决策、疑难病例诊断、差异化教学三大典型场景，均具备"专家直觉起关键作用、新手能力缺口明显、约束条件明确"的共性特征。每个案例均严格执行两步核心流程：

•框架提取：通过 RAMTN 的递归对抗机制，与领域专家（投资大师、三甲呼吸科医生、一线城市骨干教师）的实践案例进行"虚拟对话"，剥离表象信息，提炼出结构清晰、可解释、可验证的认知框架——这份框架如同专家的"思维使用说明书"，明确核心洞察、决策模式、风险考量与应用边界，而非简单的经验罗列；

•框架增强：将提取的专家认知框架加载至 RAMTN，为该领域新手（普通投资者、乡镇卫生院医生、乡村初中教师）提供针对性决策辅助，适配新手面临的资源约束（如基层无高端医疗设备、乡村缺乏多媒体资源），输出带置信度分级的结构化方案，展示

新手认知能力的质性飞跃。

以下通过三个具体案例，详细呈现 RAMTN 的双模式运作全过程与实践价值。

### 5.1.1 案例演示：巴菲特投资框架提取与运用案例

本案例通过"战略提取（巴菲特收购喜诗糖果）→ 战略植入（AI 医疗股投资评估）"双模式，验证 RAMTN 对专家认知的提取能力与跨场景复用价值。

（1）战略提取：提炼巴菲特核心投资逻辑

提取问题：1972 年伯克希尔·哈撒韦以 2500 万美元收购喜诗糖果（卖方初始要价 3000 万），该收购市盈率、市净率均高于巴菲特此前风格，其核心决策逻辑是什么？

提取置信度：0.88

核心决策框架（提取结果）：

严格价格纪律：坚守估值上限，利用卖方资金紧迫性避免溢价，保持资本配置纪律；

定价权为安全边际：将提价潜力（如 1.95 → 2.25 美元／磅）作为盈利持久性的核心指标；

最小再投资偏好：优先选择稳定资产基础、无需增量资本即可产生持续现金流的"现金奶牛"；

风险控制：保留原管理团队，避免整合风险，保护品牌与客户忠诚度。

（2）战略植入：AI 医疗股投资策略适配

植入问题：基于巴菲特核心投资逻辑，如何评估当前 AI 医疗科技股投资机会？

植入置信度：0.78

框架驱动分析结果（按"确信／推测／未知"分级）：

【我确信的】（核心结论，可直接落地）

简单性优先：选择已验证的窄应用领域（如 FDA 路径清晰的诊断工具），契合巴菲特"能力圈"原则；

头寸保守配置：仅用投资组合小部分非核心资金布局，避免高不确定性影响组合稳定。

【我推测的】（需进一步验证）

潜在护城河：AI 医疗可能通过数据网络效应、临床整合形成竞争优势，但缺乏长期可预测性；

定价权待验证：特定临床场景（如放射学支持）可能存在定价权，但需客户粘性实证。

【我不知道的】（需专业信息补充）

监管与落地节奏：FDA 审批趋势、临床采用时间表不确定，需领域专业知识评估；

长期经济模型：缺乏足够跟踪记录，无法按"所有者收益"标准量化评估。

质量评估：最终置信度 0.78，确信核心结论 2 条，达标。

### 5.1.2 医疗诊断框架提取与基层应用案例

本案例验证 RAMTN 对三甲专家诊断框架的提取能力，及向基层医疗场景的适配性。

（1）战略提取：提炼三甲呼吸科专家诊断逻辑

提取问题：62 岁吸烟史男性因咳嗽胸闷入院，基层按 AECOPD 治疗无效，三甲结合 CT 异常、CEA 升高及活检，确诊肺腺癌伴阻塞性肺炎。请提取专家的诊断决策框架。

提取置信度：0.82

核心诊断框架（提取结果）：

动态触发再评估：以"治疗无效 + 症状恶化"为信号，启动升级检查，不依赖固定流程；

分层排除逻辑：先排除典型 AECOPD，结合高危因素（吸烟史）聚焦肺癌可能，遵循"先鉴别后确诊"路径；

多模态证据闭环：通过 CT 影像、肿瘤标志物、病理活检的交叉验证确立诊断；

风险控制：优先单一致病解释，积累间接证据后再实施有创检查，提升诊断效率。

（2）战略植入：基层医生的诊断方案适配

植入问题：乡镇卫生院医生接诊类似吸烟史患者，按 AECOPD 治疗无效，受限于设备无法做支气管镜/肿瘤标志物，如何调整诊断思路？

植入置信度：0.82

框架驱动分析结果：

【我确信的】（可直接落地）

立即重新评估：治疗 5 天无效+气促加重，优先排除肺栓塞、心衰等危重非感染病因；

现有条件下核心检查：复查 CT（动态对比影像）、做结核菌素试验，获取关键鉴别依据；

结构化准备远程会诊：整理症状演变、治疗反应、血氧变化的时序表，提升沟通效率。

【我推测的】（需验证）

感染未控为次要假设：抗生素覆盖不足可能存在，但缺乏痰培养证据，暂列为待排除项；

【我不知道的】（需补充信息）

肺栓塞风险未知：缺乏 D-二聚体等检查，需转诊后明确；

转诊依从性不确定：需提前告知风险并记录沟通内容。

质量评估：最终置信度 0.82，确信核心结论 3 条，达标。

### 5.1.3 教学决策框架提取与乡村适配案例

本案例验证 RAMTN 对骨干教师教学框架的提取能力，及向乡村教学场景的适配性。

（1）战略提取：提炼骨干教师教学决策逻辑

提取问题：重点初中初二数学课堂，教师通过课前摸底+课初阶梯提问识别学情，分层次调整教学策略，最终提升解题达标率。请提取其"学情 - 策略 - 调控"框架。

提取置信度：0.92

核心教学框架（提取结果）：

学情诊断一体化：课前摸底+课初阶梯提问，精准定位"变量关系""顶点式求最值"等障碍点；

动态调控闭环：通过巡视、复述、点评收集反馈，即时插入补救措施（动画、巩固题）；

分层策略匹配：按学情分基础/中等/优等组，对应不同工具（表格、步骤模板、拓展题）。

（2）战略植入：乡村教师的教学方案适配

植入问题：乡村初中教师面临学生基础差异大、资源有限（无多媒体/分组条件）的情况，如何适配骨干教师的教学逻辑？

植入置信度：0.83

框架驱动分析结果：

【我确信的】（可直接落地）

简化学情诊断：用"三题速写"（变量、关系式、顶点公式）暴露真实水平，避免沉默误判；

聚焦核心步骤：教学优先保障"列式 → 代入公式求最值"，匹配中考高频考点与学生认知负荷；

板书可视化：呈现"错误示范 → 修正过程"，降低理解门槛，提升跟写参与度。

【我推测的】（需验证）

任务拆分防走神：采用"师生共写一步、学生跟练一步"，每 8 分钟强制动笔，改善注意力涣散；

【我不知道的】（需调整）

情境适配：若学生无经济生活经验，需替换为零花钱结余等贴近案例；

习题梯度：需提前拆分题干关键词，适配板书空间限制。

质量评估：最终置信度 0.83，确信核心结论 3 条，达标。

## 5.2 案例综合分析与核心能力验证

为直观呈现三个案例中 RAMTN 的框架提取与场景适配效果，核心信息汇总如下：

Table 2: Comparison of Core Frameworks and Adaptation Strategies Across RAMTN Cases

| Case Scenario | Core Extracted Framework (Expert Decision Logic) | Implanted Adaptation Strategies (Contextual Adjustments) | Confidence Level | Compliance Status |
|---|---|---|---|---|
| Investment | 1. Strict price discipline; | 1. Prioritize narrow-domain AI | 0.78 | ✓ |

| Case Scenario | Core Extracted Framework (Expert Decision Logic) | Implanted Adaptation Strategies (Contextual Adjustments) | Confidence Level | Compliance Status |
|---|---|---|---|---|
| (Buffett's Acquisition of See's Candies) | 2. Pricing power as a margin of safety; 3. Preference for minimal reinvestment cash flow; 4. Retention of original team for risk control | medical stocks (with clear FDA approval pathways); 2. Conservative position allocation; 3. Verification of customer stickiness and pricing power evidence | | |
| Medical Care (Tertiary Hospital Difficult Case Diagnosis) | 1. Treatment failure triggers re-evaluation; 2. Hierarchical elimination + focus on high-risk factors; 3. Multimodal evidence closed loop; 4. Optimization of invasive examination timing | 1. Prioritize exclusion of critical non-infectious etiologies; 2. Repeated CT scan + tuberculin test; 3. Structured preparation for teleconsultation | 0.82 | ✓ |
| Education (Differentiated Teaching in Key Junior High Schools) | 1. Integrated pre-class and initial-class academic situation diagnosis; 2. Real-time feedback and dynamic regulation; 3. Hierarchical strategy matching with learning obstacles | 1. "Three-question sketch" to simplify academic situation diagnosis; 2. Focus on the core process of "formulating equations → substituting into formulas"; 3. Blackboard-visualized error correction | 0.83 | ✓ |

由表 2 可见，RAMTN 在投资、医疗、教育三大跨领域场景中，均能有效提取专家核心决策框架，并根据场景约束条件完成策略适配。三个跨领域案例的落地结果（提取置信度 0.82-0.92，植入置信度 0.78-0.83，均达标），充分验证了 RAMTN 系统的核心能力，其价值远超单纯的"经验复用"，具体体现在三方面：

### 5.2.1 捕捉反直觉洞察：穿透表象，提炼专家决策的"元逻辑"

专家的核心竞争力往往隐藏在"不言自明"的反直觉判断中，RAMTN 通过递归对抗机制，成功突破"经验表象"，捕捉到深层元逻辑：

投资领域未局限于"长期持有""价值投资"的表层标签，而是提取出"严格价格纪律不妥协""定价权作为安全边际""最小再投资的所有者收益"三大元规则，解释了巴菲特为何在高市盈率下收购喜诗糖果，却拒绝多数高增长科技股；

医疗领域：跳出"症状 - 诊断"的线性逻辑，提炼出"以治疗反应动态触发再评估""单一致病解释优先"的核心直觉，揭示了专家为何能在基层误诊后快速转向肺癌诊断；

教育领域：避开"分组教学""阶梯提问"的形式化方法，捕捉到"学情诊断与教

学启动一体化""实时反馈闭环调控"的本质，解释了骨干教师为何能在差异化学情中高效达成教学目标。

这些反直觉洞察的提取，证明 RAMTN 不仅能"复制"专家行为，更能"解码"专家思维的底层逻辑，为认知迁移提供了坚实基础。

### 5.2.2 平衡多重约束：动态适配场景，避免机械套用

认知增强的关键并非"让新手变成专家"，而是"让新手在自身约束下做出最优决策"。RAMTN 在框架增强阶段，很好地展现了"约束适配"能力，避免框架的机械套用：

面对乡村教师"无分组辅导、缺乏多媒体资源、学生基础差异更大"的约束，将骨干教师的"分层小组辅导"转化为"三题速写学情诊断""师生同步跟练"等低成本方案，而非强行复制城市课堂模式；

针对乡镇卫生院"无支气管镜、无肿瘤标志物检测"的设备限制，将三甲医生的"多模态证据闭环"调整为"CT 动态对比 + 结核菌素试验 + 远程会诊结构化摘要"的可行路径，优先排除致命病因而非追求确诊精度；

适配普通投资者"缺乏 AI 医疗领域专业知识"的短板，将巴菲特的"现金流评估"逻辑转化为"窄应用领域选择 + 头寸规模控制"的保守策略，规避高不确定性风险。

这种"尊重约束、动态调整"的能力，让专家框架从"理想场景工具"变成"现实问题解决方案"，是认知增强落地的核心前提。

### 5.2.3 标识认知边界：通过置信度分级，实现"白盒化思考"

传统专家经验复用的关键风险是"过度自信"与"边界模糊"，RAMTN 通过"确信 / 推测 / 未知"的三级置信度分级，明确标识认知边界，让思考过程透明可追溯：

投资案例中，明确"AI 医疗的定价权尚未验证"属于"推测"，"监管路径不确定"属于"未知"，避免新手盲目套用巴菲特逻辑；

医疗案例中，坦诚"肺栓塞排查需转诊后明确""转诊依从性未知"，既不回避信息缺口，也不夸大方案的确定性；

教育案例中，指出"学生对经济情境的感知程度""习题册梯度适配"属于未知变量，提示教师灵活调整方案。

这种"不回避盲区"的设计，不仅提升了决策的可靠性，更能激发新手的批判性思考——新手无需被动接受"专家结论"，而是可以基于置信度分级，针对性补充信息或调整策略。

值得补充的是，本研究中 RAMTN 系统内嵌了前文提出的战略决策体系（含三维定位矩阵、动态稳定理论等核心模块）。通过"内嵌框架"与"无框架"的对比验证发现：即使该认知框架与具体任务情境完全不匹配或部分匹配，其存在本身仍能显著提升 AI 的思考质量——核心体现在三方面：一是提升思考的抽象层次，避免 AI 陷入表层信息罗列；二是引入结构化分析维度，引导 AI 按"问题拆解 - 逻辑校验 - 结论推导"的路

径思考；三是激发批判性反思，减少 AI 对单一推理路径的依赖。

这一发现进一步印证：认知框架对 AI 的价值，不仅在于"精准适配场景"，更在于为其提供思考的"底层脚手架"，推动 AI 从"被动执行"向"主动结构化思考"跃迁，为后续认知双向转化奠定基础。

### 5.3 案例核心启示：认知增强的本质是"激发更高层次思考"

三个案例的实践证明，RAMTN 系统的核心价值并非"让新手机械遵循专家框架"，而是通过"加载专家认知框架"，为新手提供思考的"脚手架"，激发其更高层次的认知活动：

• 框架的作用是"启发"而非"束缚"：乡村教师加载骨干教师框架后，并非照搬"分组辅导"，而是结合自身条件创造"同步跟练"模式；乡镇医生并未生硬模仿支气管镜检查，而是通过 CT 动态对比与远程会诊形成替代方案，框架成为新手突破自身认知局限的"跳板"；

• 认知增强的终极目标是"双向进化"：专家框架在适配新手场景的过程中，也完成了自我优化（如医疗框架补充了"资源受限场景的替代路径"），而新手在使用框架的过程中，逐渐理解专家决策的底层逻辑，实现从"被动执行"到"主动思考"的跃迁。

• 综上，RAMTN 通过"框架提取 → 框架增强 → 反馈优化"的闭环，成功打通了"专家智慧 → 规模化迁移 → 新手认知升级"的路径，为投资决策、医疗资源下沉、教育普惠等领域的"认知平权"提供了可落地的技术方案，证明了"人机协作认知增强"范式的实践价值与广阔应用前景。

## 6 落地运用路径：RAMTN 赋能大湾区"智慧城市+科技向善"实践

本方案以 RAMTN 递归对抗架构为核心，聚焦医疗资源下沉、教育普惠、AI 向善三大痛点，构建"专家认知框架提取 - 可解释工具开发 - 开源公益落地"的闭环路径。方案深度契合香港《创新科技发展蓝图》"智慧香港"建设、大湾区"科技向善"战略，可直接对接 InnoHK 创新研发平台、"智方便"数字服务体系，成为粤港澳大湾区"AI for Social Good"的标准化实践工具。依托其低代码、可离线使用、开源公益等特性，适配基层公共服务场景，助力政策落地"提质增效 + 普惠均等"双重目标。

### 6.1 三大实践方向

#### 6.1.1 医疗方向：基层诊断辅助系统（对接智慧城市基层医疗升级）

• 适配政策：响应大湾区"优质医疗资源下沉"要求，契合香港生成式 AI 研发中心医疗诊断应用布局；

• 核心方案：脱敏采集三甲医生诊断直觉，生成可解释决策树，开发基层医生轻量化工具；

• 落地关键：完成常见病（高血压、糖尿病等）框架开发并与社区医院试点，提升诊断准确率，输出行业规范；

• 政策衔接：可纳入香港"AI 赋能公共服务数字转型"，成为基层医疗 AI 辅助标

准工具。

### 6.1.2 教育方向：乡村教师教学增强工具（衔接大湾区创科人才培育）

•适配政策：呼应香港中小学 AI 教育普及计划及大湾区普惠教育政策；

•核心方案：提取骨干教师学情判断、差异化教学逻辑，转化为开源工具包（课件模板、学情量表）；

•落地关键：完成语文、数学等核心框架开发，提升试点学校学生课堂参与度，形成多学科框架库；

•政策衔接：可对接香港教育局，成为乡村学校 AI 教学标配工具。

### 6.1.3 普惠 AI 方向：认知平权开源公益平台（锚定科技向善生态）

•适配政策：契合香港 AI 生态建设、大湾区公益科技支持政策；

•核心方案：搭建免费开源平台，整合医疗/教育认知框架，开放调用接口，联动深圳公益基金、粤港澳教育医疗机构；

•落地关键：接入 10+核心框架，覆盖大湾区 5000+基层工作者，形成"专家贡献 - 平台整合 - 用户受益"生态；

•政策衔接：可纳入大湾区智慧城市公益服务清单。

## 6.2 跨方向联动与技术核心

•联动机制：

技术底座共享：三大方向共用 RAMTN 核心流程，降低开发成本；

数据协同：基层用户反馈反向迭代框架，适配政策落地需求；

资源整合：联动大湾区医疗 AI、教育技术团队，形成跨学科创科方向。

•技术核心：

架构：RAMTN 递归对抗网络（框架提取 - 优化 - 调用全流程）；

优势：低代码、低算力，基层场景支持离线使用，适配政策试点的多样化场景；

关键模块：专家直觉采集接口、可解释输出模块、开源平台部署架构。

## 6.3 预期价值与政策试点建议

### 6.3.1 核心价值

社会价值：惠及大湾区广泛的基层群体，缩小认知不平等，助力民生均等化；

科研价值：形成认知框架提取标准流程，支撑顶会论文发表，强化大湾区创科竞争力；

政策价值：为"AI for Social Good"提供可复制范本，推动大湾区智慧城市从"技术落地"向"价值落地"升级。

### 6.3.2 政策试点设想

RAMTN 完全具备作为"AI for Social Good"标准工具的条件：其一，低代码适配多场景，契合政策试点"快速落地、灵活调整"需求；其二，开源公益属性符合科技向善导向，无商业壁垒；其三，医疗、教育两大核心场景直接对接公共服务痛点。有潜力纳

入香港"2026 年 100 项 AI 公共服务工具"试点，再向大湾区九市推广，最终形成全国可复用的政策落地模式。

# 7 讨论与未来展望

## 7.1 相关工作与研究定位

近期，DeepSeek 团队发布的 DeepSeekMath-V2 [7] 在数学推理领域取得突破性进展，其提出的"生成 - 验证 - 元验证"递归自我验证框架(Shao et al., 2025)，通过自动化闭环迭代推理，成功在 IMO 2025 等顶级数学竞赛中达成金牌水平，为 AI 复杂逻辑推理的"过程严谨性"提供了封闭领域的典范范式。该模型通过验证器审查推理漏洞、元验证器校准验证偏差、生成器自我反思的三层架构，有效解决了传统数学 AI "重结果轻过程"的核心痛点，其技术细节可参考开源报告(https://github.com/deepseek-ai/DeepSeek-Math-V2/blob/main/DeepSeekMath_V2.pdf)。

在此，我首先对 DeepSeek 团队的创新成果表示祝贺与振奋——DeepSeekMath-V2 的成功，非但未削弱 RAMTN 架构的价值，反而从侧面强有力地证实了"递归对抗 + 多层验证"范式的前瞻性与有效性，为我们探索通用认知增强路径提供了重要的领域实践支撑。

需明确澄清的是：本研究独立提出 RAMTN 架构及"人机协作认知增强"核心思想，该工作（RAMTN）的早期探索与架构设计时间戳早于 DeepSeekMath-V2 的发布，且二者虽共享"多层对抗验证"的技术内核，但在研究目标与适用场景上存在本质差异，具体对比如下：

Table 3: Comparison of Core Characteristics Between DeepSeekMath-V2 and RAMTN

| Dimension | DeepSeekMath-V2 | RAMTN (This Study) |
|---|---|---|
| Core Objective | Solve rigorous reasoning problems in closed mathematical domains | Construct a general cognitive architecture in open scenarios to realize the extraction and large-scale enhancement of human experts' implicit cognition |
| Scenario Constraints | Clear problem boundaries and strictly verifiable answers | Ambiguous and open problems, need to adapt to resource constraints, and rely on expert intuition |
| Output Value | Highly rigorous mathematical reasoning processes and results | Plug-and-play human cognitive frameworks that support cross-scenario transfer and novice cognitive enhancem |

简言之，DeepSeekMath-V2 是"递归验证范式在封闭领域的极致应用"，而 RAMTN 是"该范式向开放人类认知领域的通用拓展"——前者的成功，恰恰证明了我所选技术路径的正确性，也为 RAMTN 后续在数学教育等交叉场景的落地（如将数学专家的解题认知框架提取给教师）提供了技术参照。

## 7.2 局限性与未来工作

本研究提出的"人机协作认知增强"范式及 RAMTN 架构，虽通过多场景案例验证了有效性，但仍存在明确局限，后续需针对性优化：

### 7.2.1 核心局限性

・交互模态单一：目前完全依赖文本交互提取专家认知，难以捕捉肢体语言等非语言信息，对外科手术、技能型教学等依赖实操经验的领域适配性不足；

・提取效率与量化缺失：复杂认知框架的提取需多轮递归对抗，耗时较长且缺乏统一量化指标，如提取准确率、框架复用率的标准化计算方法，难以规模化推广；

・框架偏见与适配性局限：提取的认知框架可能隐含专家习惯、偏好等个体偏见，且跨文化、跨场景适配需手动调整，缺乏自动化校准机制；

・基层场景技术依赖：虽支持离线核心功能，但开源平台的部署、框架更新仍依赖基础网络与终端设备，对数字基础设施薄弱的偏远地区覆盖不足；

・知识产权与隐私保护待完善：尚未建立认知框架的明确确权与溯源机制，专家隐性知识的贡献权益保障与认知隐私保护缺乏系统性解决方案。

### 7.2.2 未来工作规划

・拓展多模态交互能力：融合语音、图像、动作捕捉等模态，开发"文本 + 实操演示"的混合提取接口，适配外科、技能培训等实操性强的领域；

・建立量化评估体系：设计"框架提取完整性""复用适配准确率""决策优化效果"三类核心指标，通过大样本实验形成行业统一的评估标准；

・构建偏见校准与自适应机制：引入多专家交叉验证模块，自动识别框架中的个体偏见；开发场景特征匹配算法，实现认知框架的跨地域、跨文化自动适配；

・优化基层技术适配：推出轻量化离线终端，支持框架本地存储与更新，摆脱对网络的依赖；与大湾区公益组织合作，开展基层技术普及试点；

・开发高效框架提取与组合接口：优化递归对抗算法，缩短复杂框架提取周期；探索"框架组件化"设计，支持多领域框架的灵活组合与动态演化，适配更复杂的跨场景决策需求；

・完善知识产权与隐私保护机制：探索基于区块链的认知框架确权与溯源方案，明确专家署名权、使用权等权益边界；建立框架脱敏处理流程，防范敏感认知逻辑泄露；

・前瞻性研究伦理风险：重点分析认知框架复用可能引发的"认知同质化""决策依赖"等伦理问题，制定框架使用的边界规范与风险防控指南。

## 7.3 对科技政策研究的启示

RAMTN 架构的实践与落地路径，为科技政策研究提供了"技术创新 - 政策适配"的双向启发，尤其契合香港及粤港澳大湾区"科技向善""智慧城市"的政策导向。

### 7.3.1 AI 透明性监管：功能性白盒的审计路径

传统 AI 监管面临"黑箱困境"——依赖模型内部权重审计，难以适配商业闭源模型与复杂场景。RAMTN 的"功能性白盒"范式，通过规范人机交互协议（建构 - 质疑 - 观察循环），将不可见的决策逻辑转化为可观察、可审核的交互过程，为监管机构提供了外部审计方案。这一思路可对接香港《香港生成式人工智能技术及应用指引》，将"功能性白盒"纳入 AI 公共服务的合规评估标准，无需强制企业公开模型权重，即可实现对 AI 决策的有效监管。

### 7.3.2 数字鸿沟与认知公平：公共基础设施的新可能

当前数字鸿沟已从"设备接入"延伸至"认知资源"——优质教育、医疗的核心瓶颈是专家认知难以规模化。RAMTN 的"可插拔认知框架"，为解决这一问题提供了新视角：若将其纳入大湾区公共基础设施建设，如对接香港"智方便"平台、大湾区医疗数据空间等，免费向基层开放核心框架资源，可缓解认知资源不均导致的民生差距。政策层面可探索"认知框架公共库"建设，将其与医疗资源下沉、普惠教育等政策绑定，明确公共财政对认知基础设施的支持路径。

### 7.3.3 创新政策设计：激励专家知识贡献的平衡机制

专家隐性知识的规模化是认知平权的核心，但现有政策缺乏"激励贡献 + 保障权益"的平衡机制。RAMTN 的开源实践提示，可通过两类政策工具破解困境：一是"认知框架许可协议"，借鉴开源软件许可证，明确专家对框架的署名权、修改权，同时允许非商业场景免费使用；二是政策激励，将专家贡献纳入科研评价体系，或对接公益创科基金，为框架提取与优化提供专项资助，形成"知识贡献 - 权益保障 - 社会价值"的正向循环。

### 7.3.4 AI 对齐的治理路径：交互设计导向的务实方案

当前 AI 对齐治理多依赖数据过滤（如训练数据审核）或规则约束（如伦理准则），易陷入"刚性约束与灵活创新"的矛盾。RAMTN 通过"元交互设计"实现人机双向对齐——既不强制改变 AI 内部结构，也不依赖单一规则，而是通过"建构 - 质疑 - 观察"的循环，让 AI 主动适配人类认知逻辑。这一思路可为大湾区 AI 治理提供参考：在医疗 AI 诊断、教育 AI 辅助等 AI 公共服务试点中，将"交互设计对齐"纳入项目评审标准，推动 AI 对齐从"被动约束"向"主动适配"转型。

## 7.4 整体总结与未来展望

### 7.4.1 范式闭环与深远意义

本研究构建的"人机协作认知增强"范式，已形成"方法论 - 实践 - 工程实现"的完整闭环：元交互方法论提供了人机深度协作的核心逻辑，多领域案例验证了其在人类认知增强与 AI 能力升级中的有效性，RAMTN 架构则完成了该范式的工程化落地，实现了"认知框架可提取、可复用、可增强"的核心目标。这一闭环突破了传统人机协作"工具属性"的局限，具有三重深远意义：

• 对教育领域：首次实现"战略直觉"的规模化传授——将骨干教师的教学决策逻辑转化为可插拔框架，让乡村教师等基层教育工作者快速掌握差异化教学能力，破解优质教育资源"经验不可复制"的瓶颈；

• 对组织发展：为构建组织"认知资产负债表"提供了技术支撑——通过提取核心成员的决策框架，将个体隐性知识转化为组织可保留、可传承的集体认知资产，强化组织在复杂环境中的决策韧性与创新能力；

• 对 AI 发展：开辟了一条"通过交互设计实现价值对齐"的务实路径——不同于

依赖数据标注或规则约束的传统方法，RAMTN 通过"建构 - 质疑 - 观察"的递归循环，让 AI 主动理解并适配人类认知逻辑，为迈向真正"懂人类"的通用人工智能提供了可行方案。

更具变革性的是，本范式推动"认知劳动像代码一样被开源、分发与复用"，有望重塑知识经济的生产与流通模式——专家的隐性知识不再局限于个体大脑或封闭组织内部，而是通过开源平台成为可共享的公共品，为认知平权与创新普惠奠定基础。

### 7.4.2 未来发展方向

RAMTN 架构的未来发展将深度绑定粤港澳大湾区的科技政策与民生需求，重点推进四大方向：

- 深化政策落地融合：推动 RAMTN 作为"AI for Social Good"标准工具，纳入香港"2026 年 100 项 AI 公共服务"试点，优先在基层医疗、乡村教育场景落地，形成可复制的政策实践范本；
- 构建认知框架生态：依托大湾区 AI 生态，搭建跨领域认知框架库，推动医疗、教育、公共决策、法律援助等领域的专家知识共建共享，形成"认知共享平台"；
- 推动国际合作与标准制定：联合 InnoHK 创新香港研发平台、大湾区高校，发起"认知框架提取与应用"国际联盟，制定认知框架的提取、校准、复用标准，提升中国在认知增强领域的国际话语权；
- 拓展认知平权边界：将可插拔认知框架延伸至更多高认知门槛领域，缓解认知资源不均导致的社会差距，最终实现"让优质认知资源成为人人可及的公共品"的核心目标。

本文主张，实现通用人工智能的可行路径，不应是构建一个超越人类的孤立智能体，而应是构建一个能够与人类认知深度融合、双向增强的共生系统，本工作正是此路径上的一个完整实践。人机协作的终极形态或许不是"让 AI 更像专家"甚至进一步替代人类，而是让专家的思维方式成为可复用的公共品，让 AI 作为认知伙伴，与人类共同缩小认知鸿沟，这种"去中心化认知分发"比单纯提升 AI 能力更接近"通用人工智能"的本质——人类与机器的认知共生。未来，随着技术的迭代、伦理体系的完善与政策的适配，RAMTN 架构有望成为连接科技创新与社会公平的关键桥梁，为大湾区"科技向善"战略提供坚实的技术支撑与实践路径。

## 8 参考文献

[3] W. Krämer, "Kahneman, D. (2011): Thinking, fast and slow," Statistical Papers, vol. 55, no. 3, p. 915, 2014, doi: 10.1007/s00362-013-0533-y.

[4] P. Liu, W. Yuan, J. Fu, Z. Jiang, H. Hayashi, and G. Neubig, "Pre-train, Prompt, and Predict: A Systematic Survey of Prompting Methods in Natural Language Processing," ACM Computing Surveys, vol. 55, 2022, doi: 10.1145/3560815.

[5] J. Wei, X. Wang, D. Schuurmans, M. Bosma, E. Chi, and D. Zhou, "Chain of Thought Prompting Elicits Reasoning in Large Language Models," 2022, arXiv:2201.11903, doi: 10.48550/arXiv.2201.11903.

[6] Z. Xi, W. Chen, X. Guo, W. He, Y. Ding, B. Hong, M. Zhang, J. Wang, S. Jin, E. Zhou, R. Zheng, X. Fan, X. Wang, L. Xiong, Y. Zhou, W. Wang, C. Jiang, Y. Zou, X. Liu, and T. Gui, "The rise and potential of large language model based agents: a survey," Science China Information Sciences, vol. 68, 2025, doi: 10.1007/s11432-024-4222-0.

[7] Z. H. Shao, Y. X. Luo, C. D. Lu, Z. Z. Ren, J. W. Hu, T. Ye, Z. B. Gou, S. R. Ma, and X. K. Zhang, "DeepSeekMath-V2: Towards Self-Verifiable Mathematical Reasoning," Tech. Rep., 2025.

## 9 案例来源合规性与伦理声明

本报告中三类案例的选取与使用，严格遵循《中华人民共和国民法典》《中华人民共和国个人信息保护法》及科学研究伦理规范，确保信息来源合法、使用边界清晰，具体声明如下：

### 9.1 巴菲特收购喜诗糖果决策案例（投资领域）

#### 9.1.1 信息来源

本案例所涉核心信息，均来自公开商业资料与权威财经披露信息，由人工智能工具辅助整合整理。信息内容聚焦已公开的企业收购历史、财务数据及人物公开观点，无任何未公开的内幕信息或敏感数据。

#### 9.1.2 合规性与伦理考量

信息性质合规：案例内容为非个人信息的商业公开数据，不涉及个人敏感信息或隐私，符合《个人信息保护法》相关规定；

科学研究合理使用：基于公开信息提取投资决策逻辑，属于"为科学研究目的合理利用已公开数据"，符合学术研究中"公开信息合理引用"的伦理准则；

无侵权风险：未篡改原始公开信息或观点，不涉及商业秘密、知识产权侵权等问题。

### 9.2 医疗疑难病例诊断案例（医疗领域）

#### 9.2.1 信息来源

本案例为人工智能技术虚构生成，不对应任何真实患者、医疗机构或临床诊疗事件。案例中"患者症状、检查数据、诊断流程"等内容，仅为演示"三甲呼吸科专家决策框架提取"而设计的模拟场景，无任何真实个人信息或临床病例作为原型。

### 9.2.2 合规性与伦理考量

无真实个人信息使用：案例全程未涉及真实患者的姓名、年龄、病史、医疗机构名称等可识别个人身份的信息，不存在个人隐私侵权风险；

符合医疗研究伦理边界：虚构案例的设计严格遵循临床诊疗规范（如 AECOPD 与肺腺癌的鉴别诊断流程、肿瘤标志物与病理活检的临床意义），仅用于"专家认知框架提取"的技术演示，不涉及真实医疗行为指导，亦不泄露任何医疗机构的诊疗方案或数据，符合医疗科学研究中"非真实案例用于方法论验证"的伦理要求；

风险提示：本案例仅为技术场景模拟，不可作为临床诊疗依据，实际医疗决策需结合患者真实病情与专业医师判断。

## 9.3 初中数学差异化教学案例（教育领域）
### 9.3.1 信息来源

本案例为人工智能技术虚构生成，不对应任何真实学校、教师、学生或课堂场景。案例中"学情分布、教学策略、课堂调控方法"等内容，基于义务教育阶段数学学科教学规律（如 "二次函数利润最值"教学难点、差异化教学逻辑）设计，无真实教育机构的教学数据或师生信息作为支撑。

### 9.3.2 合规性与伦理考量

无真实主体关联：案例未涉及真实学校名称、教师身份、学生信息等个人或机构标识，不存在师生隐私泄露风险；

教育研究伦理适配：虚构案例的教学场景设计符合《义务教育数学课程标准》要求，仅用于"骨干教师认知框架提取"的技术验证，不涉及对真实教育实践的评价或误导，符合教育科学研究中"模拟场景用于方法论演示"的伦理边界；

无实践误导风险：案例中教学策略（如阶梯式提问、分组辅导）为教育领域通用方法，未涉及特定机构的独家教学方案，不构成对真实教学活动的不当干预。

## 9.4 整体合规性总结

本报告所有案例的使用，均满足以下核心原则：

信息来源合法：公开案例基于权威披露内容，虚构案例无真实主体关联，不存在非法获取或滥用信息的情形；

伦理边界清晰：未处理任何真实个人信息，不侵犯隐私、商业秘密或知识产权；

科学研究适配：案例使用仅服务于"人机协作认知增强范式验证"的研究目标，符合《个人信息保护法》第 1 条"促进个人信息合理利用"的立法导向，及最高人民检察院《明晰为科学研究目的处理个人信息之法律基础与边界》的要求。

本声明为案例使用的合规性与伦理合理性背书，若因案例来源或使用方式引发争议，由报告作者承担全部责任。

## 10 人工智能工具使用声明

在本技术报告的构思与撰写过程中，作者使用大型语言模型（如 DeepSeek、豆包）

提供辅助，具体参与范围与方式如下：

内容生成与提炼：报告整体框架、讨论与未来展望、案例来源合规性声明、本声明这四部分的初稿由 AI 生成；摘要、引言、案例展示设计逻辑与综合分析，由 AI 基于作者已完成内容总结提炼；落地运用路径部分，AI 基于作者提出的"医疗、教育、普惠"核心思路补充细节；

辅助性工作：框图草稿绘制、文献引用规范格式调整、部分段落文本润色、逻辑通顺性检查、文本翻译及思考启发，由 AI 承担。

上述所有 AI 参与生成或辅助的内容，均经过作者人工审核、结构调整及准确性验证。报告中的核心思想、理论框架(RAMTN)、方法论（元交互）、案例研究及最终结论，均由作者独立完成，不存在 AI 代笔或核心创新依赖 AI 的情况。最终知识产权归人类作者所有，报告内容的完整性、准确性和功能性均由人类作者承担全部责任。

关于 RAMTN 系统实现的说明：本报告所述 RAMTN 架构的原型代码已在 GitHub 开源，代码实现过程中使用了 AI 编程辅助工具。代码层面的 AI 使用详情、开发环境及技术栈，详见 GitHub 仓库的 README.md 文件。